# ENHANCING RICE LEAF IMAGES: AN OVERVIEW OF IMAGE DENOISING TECHNIQUES

Rupjyoti Chutia[1*], Dibya Jyoti Bora[1]

[1]*Department of Information Technology, School of Computing Sciences, The Assam Kaziranga University, Koraikhowa, NH-37, Jorhat 785006, Assam, India.*

*\*Corresponding Author:*
*E-mail: rupom.chutia.jrt@gmail.com*

**ABSTRACT.** The systematic and meticulous handling and processing of digital images through use of advanced computer algorithms is popularly known as the digital image processing. It has received significant attention in both academic and practical fields. Image enhancement serves as a crucial preprocessing stage in each of the image-processing chain. It enhances the quality of the image and emphasizes its features, making all subsequent tasks (segmentation, feature extraction, classification) more reliable and accurate. Image enhancement is also essential for rice leaf analysis, particularly for disease detection, nutrient deficiency evaluation, and growth analysis. Denoising followed by contrast enhancement of the images are the primary steps of image enhancement and image preprocessing. Image filters are generally employed as image denoising techniques. Image filtering operations are designed to transform or enhance the visual characteristics of an image. These include properties such as brightness, contrast, color balance, and sharpness. Thus, they play a very significant and crucial utility function in enhancement of the overall image quality and enabling the extraction of useful information from the visual data. In the current work, to provide an extensive comparative study of some of the well known image-denoising methods combined with a popular histogram equalization technique CLAHE (Contrast Limited Adaptive Histogram Equalization) for efficient denoising of rice leaves image. The experimental part of this work was performed on a rice leaf image dataset to ensure that the data used in the study is relevant and representative. The results of these experiments were then closely examined using a variety of different metrics to ensure that the image enhancement methods are tested thoroughly and comprehensively. This approach not only provides a strong basis for assessing the effectiveness of various methodologies in the digital image processing but also reveals certain insights that might be useful for adaptation in the future towards agricultural research, and other varied domains.

## INTRODUCTION

Rice is a major staple worldwide, representing a cornerstone, in terms of caloric intake, of the human diet. Correct diagnosis of rice leaf disease and nutrient deficiency is important for proper crop management, for high yield, and to minimize economic losses. With the help of Computer Vision, a sub-division of Artificial Intelligence, which is also known as digital image processing, the rice leaf images can be analyzed to provide an accurate diagnosis of disease and detect nutrient deficiency etc. Nonetheless, rice leaf images are often corrupted by different types of degradation, including Gaussian noise, Salt-and-pepper noise, and speckle noise, and so forth, which can degenerate the perceptual quality and influence the working performance of image analysis algorithms. Image noise may cause disease misclassification, wrong diagnosis and decreased accuracy in nutrient deficiency detection. Therefore, denoising and enhancement of the raw images of the rice leaves is crucial as a preprocessing step before further analysis. Thus, the effective image denoising methods should be proposed along with image enhancement techniques like histogram equalization to degenerate noise in the rice leaf image as well as maintain significant information within the leaf.

Image filters are essential tools in image processing, used to enhance, restore, and transform images by altering their properties such as size, shape, color, and smoothness. They play a crucial role in mitigating noise and improving image quality, with applications spanning various fields including photography, engineering, and digital media. Filters can be broadly categorized into linear and non-linear types, each with distinct characteristics and applications. Linear filters, such as Gaussian and mean filters are known for their speed but are less effective in noise reduction compared to non-linear filters. Non-linear filters, including median, Non-Local Means (NLM), and anisotropic diffusion filters, offer superior noise reduction but require careful parameter tuning [1].

Contrast Limited Adaptive Histogram Equalization (CLAHE) is a popular method of image enhancement, which is especially useful to enhance images having low visibility. Although CLAHE effectively improves the contrast, it is usually combined with other techniques to overcome particular challenges like noise reduction and color balance.

In this paper, a systematic comparative study on efficacy of various image denoising methods combined with CLAHE for enhancing rice leaf images is presented. During the study, CLAHE and image filters were implemented for preprocessing rice leaf images available in publicly available datasets and the results were analyzed for evaluating the efficacy of the techniques for the purpose with the help of various image quality metrics.

**IMAGE NOISE:**

Image noise refers to the random variations of pixel values unrelated to the scene that degrade image quality, often appearing in the form of a grainy overlay on an otherwise clear image. Its sources are results of some acquisition or processing operations, which should be observed properly for developing effective denoising techniques [2]. Some of the common types of noise found in images of rice leaves are Gaussian noise, Salt-and-pepper noise, Speckle noise, Random noise, and so on.

*Gaussian noise*

Gaussian noise is an additive type of noise that follows a Gaussian distribution, typically used to simulate sensor or thermal noise in images and signals [3]. It is also referred to as *white* noise. Fig. (1a) shows the original image and Fig. (1b) shows the image with Gaussian noise.

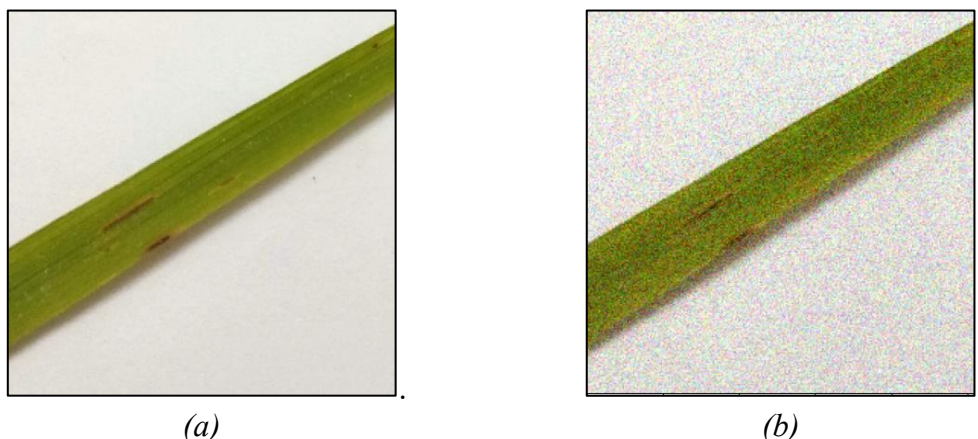

*(a)*          *(b)*
***Fig. 1.** (a) Original Image, (b) Image with Gaussian Noise*

***Salt-and-pepper noise (Impulse noise)***

Salt-and-pepper noise typically arises from errors due to malfunctions in pixel elements within camera sensors, erroneous memory locations, or timing errors during the digitization process [4]. It is an impulse type of noise where corrupted pixels are alternately altered to either their minimum or maximum values (white or black), producing a "salt and pepper" appearance in the image, while unaffected pixels retain their original values. Fig. 2(a) shows the original image and Fig. 2 (b) shows the image with Salt-and-pepper noise.

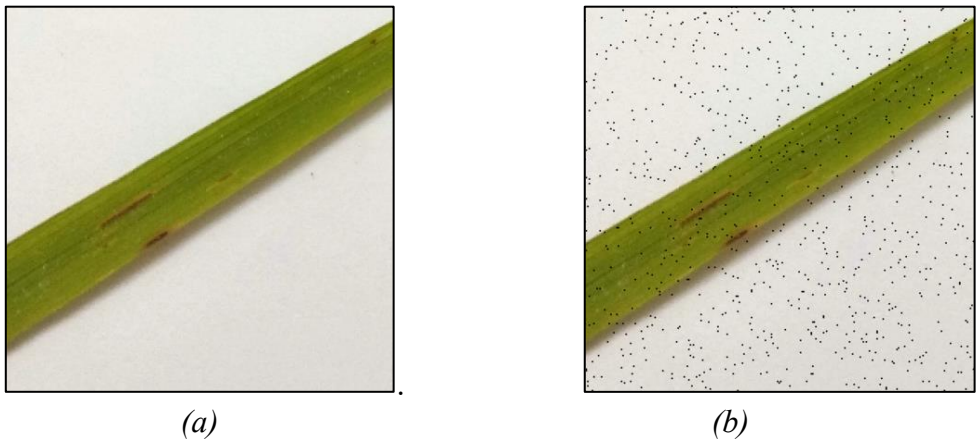

*(a)*  *(b)*
***Fig. 2.*** *(a) Original Image, (b) Image with Salt-and-pepper noise*

**Speckle noise**

Speckle noise is multiplicative granular noise due to coherent interference, generating bright and dark spot patterns that degrade contrast. Fig. (3a) shows the original image and Fig. (3b) shows the image with Speckle noise.

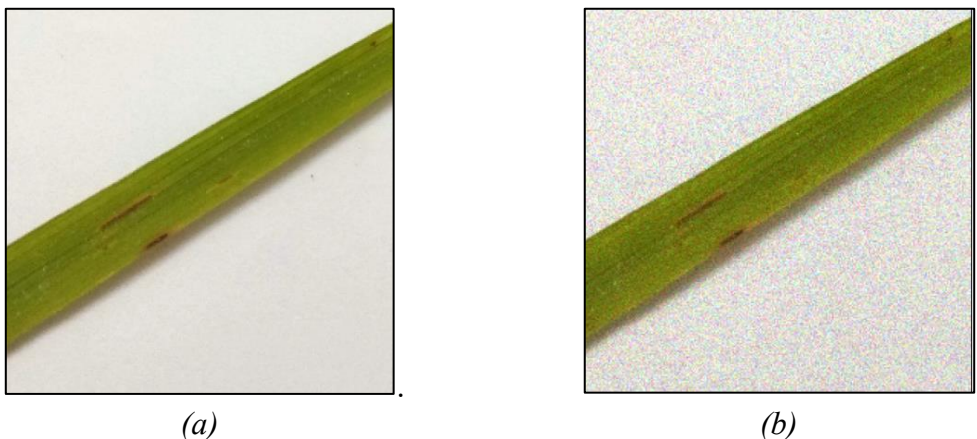

*(a)*  *(b)*
***Fig. 3.*** *(a) Original Image, (b) Image with Speckle Noise*

**Random noise**

Random noise refers to the erratic and unpredictable variations in pixel brightness or chromaticity resulting from sensor and electronic fluctuations, manifesting as grainy spots devoid of any discernible spatial arrangement. This form of noise is among the most prevalent types encountered in digital imaging devices. It is reasonable to anticipate the occurrence of a certain level of random noise, which is primarily influenced by the sensitivity parameter commonly referred to as ISO speed [5]. Fig.( 4a) shows the original image, and Fig. (4b) shows the image with Random noise.

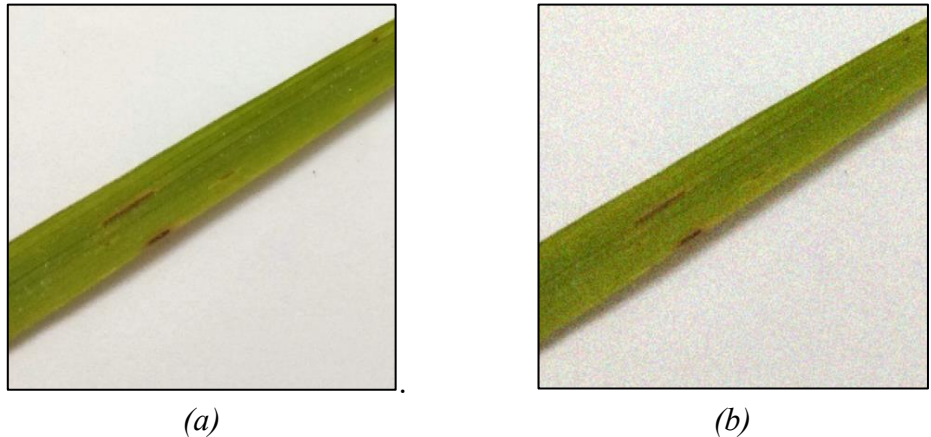

*(a)* *(b)*
***Fig. 4.** (a) Original Image, (b) Image with Random Noise*

**IMAGE ENHANCEMENT**

Image enhancement is crucial in the research of rice leaves, especially in tasks like disease diagnosis, nutrient deficiency evaluation, and growth condition surveillance. Enhanced images helps the analysis process in a various way. Few of them are:

Increased visibility of symptoms: In terms of disease or nutrient stress, subtle signs, such as discoloration, lesions, or texture changes, are hard to discern in raw images. Enhancement techniques such as CLAHE (Contrast Limited Adaptive Histogram Equalization) help clarify these values more clearly.

Noise reduction: The field images usually contain various illumination, shading or background clutter. The enhancement of the image helps in suppression of the extraneous information as well as in the accentuation of the leaf.

Standardization in Machine Reasoning: Improved images provide more consistent input for machine learning models, enhancing classification precision and mitigating false positives or negatives.

Quantitative evaluation: Several methods (e.g. histogram equalization, color normalization) are used to improve the accuracy of leaf color quantification, which is often associated with severity of nitrogen content or disease.

Automation facilitation: Such improved images significantly alleviate the tasks of segmentation, feature extraction and classification, which are the vital elements of successful development of AI-based systems for precision agriculture.

*CLAHE (Contrast Limited Adaptive Histogram Equalization):*

CLAHE is an abbreviation for Contrast Limited Adaptive Histogram Equalization- a popular method in image enhancement, especially for enhancing an image having a low contrast, noise, or inadequate illumination that finds applications in a variety of fields like medical imaging, satellite imaging and real-life photography. It works well when you have low-contrast or unevenly lit images such as rice leaves in natural fields. The method prevents the amplification of the noise by diminishing the contrast amplification inhomogeneous area of the image [6, 7].
Histogram equalization is a technique used in image processing to modify contrast by remapping intensity values such that the histogram covers the available range equally [8]. By distributing well used brightness values, it improves both global and local contrast of the image, in this way presenting the details which might be obscured. Adaptive Histogram Equalization (AHE) improves the contrast of the image by performing a histogram equalization to limited areas (tiles) of the image rather than to the whole image. This has the effect of bringing local details out. CLAHE is an improvement on AHE that restricts the contrast increase in each tile. This avoids over-enhancing noise in uniform areas (such as smooth leaf surfaces).

The two parameters of CLAHE, which specifies how the equalization is applied to the image, are:
- Clip limit: Limit the amount of contrast allowed, and
- Tile Grid Size: It defines how finely the image is divided [9].

*Image Filters*

Image filtering is an important aspect of image improvement. Filtering refers to performing mathematical operations (usually with kernels or masks) on an image in order to enhance or reduce particular features in an image.

Depending on type of filter, it may result in different behavior; it may:
- Have the image smoothed or blurred to get rid of noise.
- Make edges crisper, which highlights the boundaries and details.
- Enhance contrast or textures.
- Detect edges or specific patterns.

Image filters can be classified into two broader categories: Linear filter and Non-linear filter.

*Linear Filters*

In linear filtering, the new value for a pixel is the linear combination of surrounding pixels. The process of linear filtering can be implemented using convolutions. Mathematically, it can be expressed as:

$$O(m,n) = \sum_{i=0}^{M}\sum_{j=0}^{N} K(m-i, n-j)I(i,j) = K(m,n) ** I(m,n)$$

**Eqn. 1**

where,
- ➔ $I(m, n)$ is the input image.
- ➔ $K(m, n)$ is the filter impulse response.
- ➔ $O(m, n)$ is the output image.

Linear filtering is recognized as being one of the most effective methods of image enhancement. Simple examples are the Gaussian filter and the mean filter.

Gaussian Filter: In the field of image processing and computer vision, Gaussian filtering is a major technique with many applications. The reason it works is that images are smoothed by averaging pixel values using a Gaussian kernel. It is greatly used for noise suppression in digital images. However, while Gaussian filters are effective at reducing noise levels in images they have a downside. Smoothing out noise can cause distortion of the actual signal [10]. Therefore, if any noise is removed from an image, there is also a risk that important picture details might be blurred or lost.

Mean Filter: A mean filter is commonly employed in image processing, which is a method to reduce noise and smooth pictures by taking each pixel's value as an average of those which are near. The process greatly promotes image clarity and is essential for functions like image recognition [11]. It is effective when smoothing out random noise but can over-smooth the image and blur borders [1].

*Nonlinear Filters*

A nonlinear filter is an image processing filter in which the output signal is not a linear function of the input signal. The non-linear filter is specifically designed to use non-linearity in the signal itself. There are many applications for this particular feature, for example, noise elimination, edge enhancement in images.

Median Filter: The median filter is an order statistic filter. It examines the pixel density values of a small area within the specified filter size and selects the median intensity of the values as its center point's intensity value [12].

Bilateral filter: Bilateral filtering is one of the most sophisticated methods in image processing, according to work by Gong (2023) [13]. It can take out noise while preserving important information, especially significant edges. The nonlinear version can deal effectively with all sorts of noise, including Gaussian and salt-and-pepper type, by incorporating the spatial adjacency of pixels as well as intensity correlation.

Non-Local Means (NLM) Filter: Non-Local Means (NLM) Filter: A high-quality but computationally heavily weighted image is generated by averaging all pixels in an image, according to their resemblance [1].

- o BM3D Filter: In medical imaging, sonar, and processed compressed images, this fairly new BM3D (Block Matching and 3D Filtering) filter can be used for high-quality denoising. It works by grouping similar image patches together and doing collaborative filtering; this combination both significantly improves image quality and successfully reduces noise.

## MATERIALS AND METHODS

a) **Image acquisition:** Used 3355 images of rice leaves publicly available in Kaggle [14] for the experiment.
b) **Computing environment and programming language:** Google colab, Python.
   - Python libraries used:
     - os: File and directory operations
     - time: Time measurement and delays
     - google.colab.drive: Mounting Google Drive in Colab
     - matplotlib.pyplot: Plotting and visualization
     - numpy: Numerical array operations
     - pandas: Data manipulation with DataFrames
     - PILImage : Image input/output
     - cv2: OpenCV-based computer vision and image processing
     - scikit-image (skimage.metrics) : Computing image quality metrics (MSE, SSIM, PSNR, NRMSE, NMI)
c) **Noise incorporated:** Gaussian noise, Salt-and-pepper noise, speckle noise and random noise.
d) **Denoising techniques:** Mean, Gaussian, Median, Bilateral, and BM3D filters, were used for denoising the noisy images.
e) **Image Enhancement Algorithm:** Applied Contrast Limited Adaptive Histogram Equalization (CLAHE) to the images before or after denoising. Different parameter values for Clip Limit and Tile Grid Size was used to observe the effect of application of CLAHE.
f) **Performance evaluation:**
   Five (5) Image Quality Metrics were used for analyzing the quality of the output (enhanced) images.

   a. **MSE** (Mean Squared Error): The mean squared difference between the correspondence pixel in the reference and the test image, as a measure of overall reconstruction loss.
   b. **SSIM** (Structural Similarity Index Measure): A perceptual metric assessing the local patterns of luminance, contrast and structure in two images and computing their visual similarity.
   c. **PSNR** (Peak Signal-to-Noise Ratio): For the noisy or compressed images, the logarithmic ratio (in decibels) of the square of maximum potential pixel intensity to the MSE, representing the quality of a noisy or compressed image.
   d. **NRMSE** (Normalized Root Mean Square Error): The square root of the ratio of MSE between two sets of data or images to a reference value.
   e. **NMI** (Normalized Mutual Information) **Score**: A symmetry-invariant statistic measuring shared information between two variables or images normalized to (0,1) [15].

## RESULTS AND DISCUSSION

### Experiment Details

The images retrieved from the dataset were first resized to a uniform size of 256x256. Then, four (4) common types of noise, viz. Gaussian noise, Salt-and-pepper noise, Speckle noise, and Random noise were added to the original images using OpenCV library functions. Gaussian noise was introduced by creating a noise matrix with a given mean and variance (mean=0, variance=0.01), and adding it to the image. Salt-and-pepper noise was added by randomly flipping 5% of the pixels to 0 or 255. The original image was first multiplied by Gaussian noise, and added to the original image to simulate speckle noise. To produce an image with random noise, we added random numbers drawn from a uniform distribution in a given range (-20, +20) to the original image. One of the sample images after incorporation of noise, along with the histogram, is shown in Fig. (5a), (5b), (5c), (5d), and (5e).

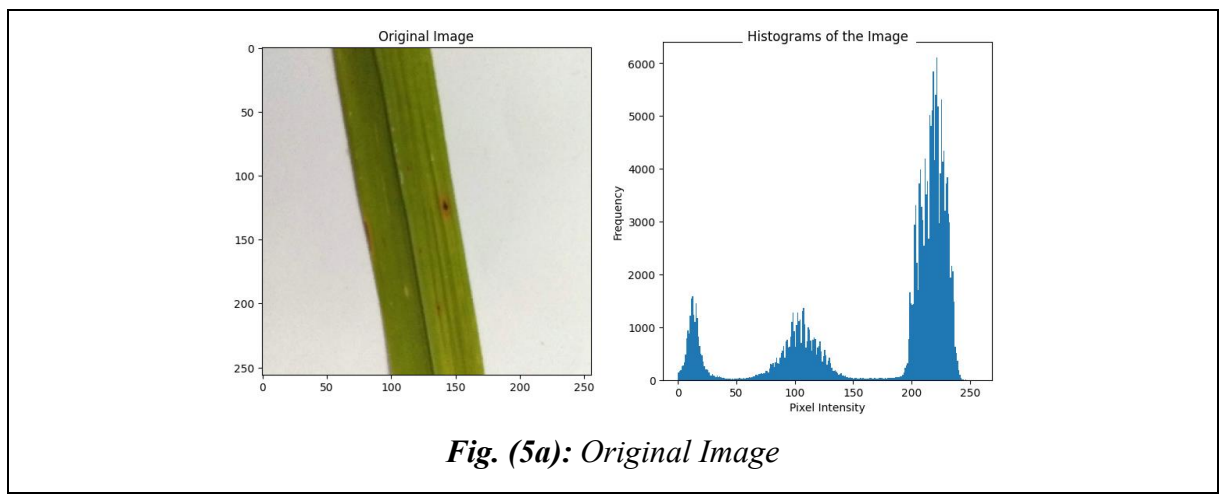

*Fig. (5a): Original Image*

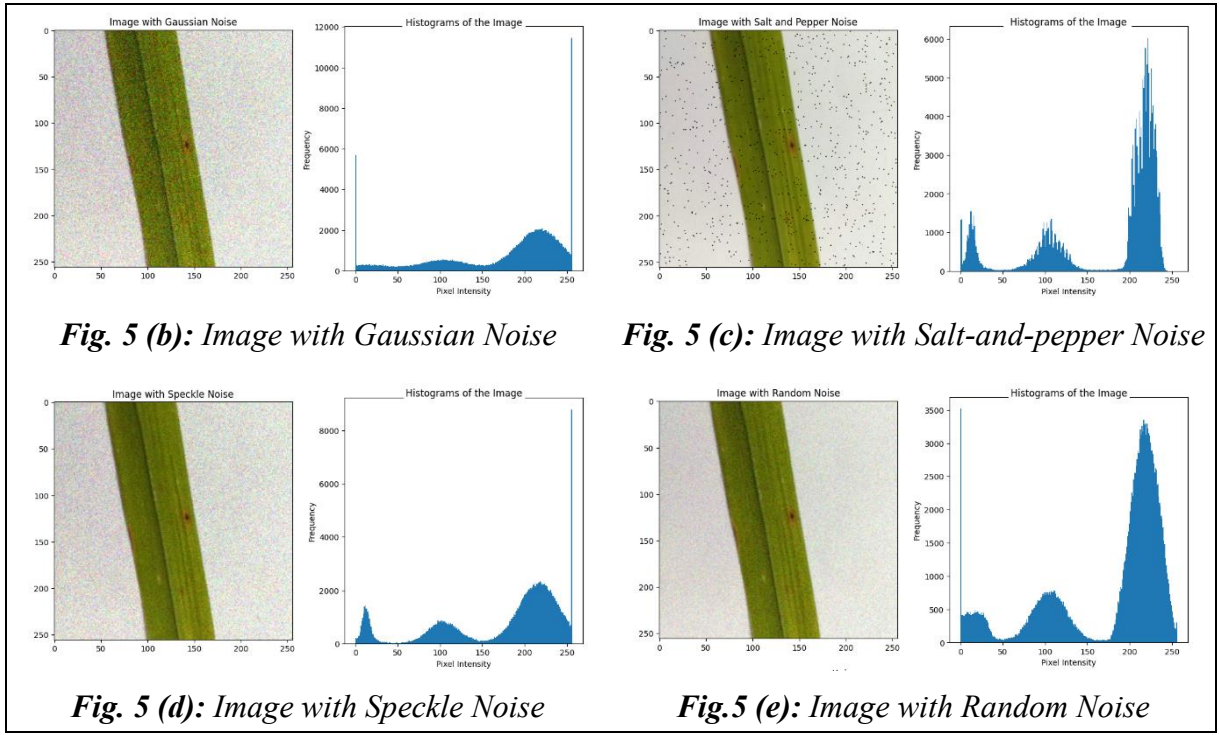

*Fig. 5 (b): Image with Gaussian Noise*  *Fig. 5 (c): Image with Salt-and-pepper Noise*

*Fig. 5 (d): Image with Speckle Noise*  *Fig.5 (e): Image with Random Noise*

Nine (9) experiments were conducted for image enhancement on the noisy images.
- A. Exp.01: Denoise
    - i. Five (5) different image filters, Mean, Gaussian, Median, Bilateral, and BM3D filters, were employed for denoising each of the noisy images.
    - ii. Analyzed the quality of the image after enhancement using image quality metrics.
- B. Exp.02: CLAHE-Denoise (2.0,8) - CD(2.0,8)
    - i. Applied CLAHE with Clip Limit=2.0 and Tile Grid Size = (8,8) to the noisy images.
    - ii. Mean, Gaussian, Median, Bilateral, and BM3D filters were applied to the result images for denoising.
    - iii. Analyzed the quality of the image after enhancement using image quality metrics.
- C. Exp.03: CLAHE-Denoise (2.0,5) - CD(2.0,5)
    - i. Applied CLAHE with Clip Limit=2.0 and Tile Grid Size = (5,5) to the noisy images.
    - ii. Mean, Gaussian, Median, Bilateral, and BM3D filters were applied to the result images for denoising.
    - iii. Analyzed the quality of the image after enhancement using image quality metrics.
- D. Exp.04: CLAHE-Denoise (1.0,5) - CD(1.0,5)
    - i. Applied CLAHE with Clip Limit=1.0 and Tile Grid Size = (5,5) to the noisy images.
    - ii. Mean, Gaussian, Median, Bilateral, and BM3D filters were applied to the result images for denoising.
    - iii. Analyzed the quality of the image after enhancement using image quality metrics.
- E. Exp.05: CLAHE-Denoise (0.5,5) - CD(0.5,5)

        i. Applied CLAHE with Clip Limit=0.5 and Tile Grid Size = (5,5) to the noisy images.
        ii. Mean, Gaussian, Median, Bilateral, and BM3D filters were applied to the result images for denoising.
        iii. Analyzed the quality of the image after enhancement using image quality metrics.

    F. Exp.06: Denoise-CLAHE (2.0,8) - DC(2.0,8)
        i. Five (5) different image filters, Mean, Gaussian, Median, Bilateral, and BM3D filters, were employed for denoising each of the noisy images.
        ii. Applied CLAHE with Clip Limit=2.0 and Tile Grid Size = (8,8) to the denoised images.
        iii. Analyzed the quality of the image after enhancement using image quality metrics.

    G. Exp.07: Denoise-CLAHE (2.0,5) – DC(2.0,5)
        i. Five (5) different image filters, Mean, Gaussian, Median, Bilateral, and BM3D filters, were employed for denoising each of the noisy images.
        ii. Applied CLAHE with Clip Limit=2.0 and Tile Grid Size = (5,5) to the denoised images.
        iii. Analyzed the quality of the image after enhancement using image quality metrics.

    H. Exp.08: Denoise-CLAHE (1.0,5) – DC(1.0,5)
        i. Five (5) different image filters, Mean, Gaussian, Median, Bilateral, and BM3D filters, were employed for denoising each of the noisy images.
        ii. Applied CLAHE with Clip Limit=1.0 and Tile Grid Size = (5,5) to the denoised images.
        iii. Analyzed the quality of the image after enhancement using image quality metrics.

    I. Exp.09: Denoise-CLAHE (0.5,5) – DC(0.5,5)
        i. Five (5) different image filters, Mean, Gaussian, Median, Bilateral, and BM3D filters, were employed for denoising each of the noisy images.
        ii. Applied CLAHE with Clip Limit=0.5 and Tile Grid Size = (5,5) to the denoised images.
        iii. Analyzed the quality of the image after enhancement using image quality metrics.

Parameter tuning of the filters: The filters were implemented using the OpenCV library functions. We conducted manual tuning of key parameters of the filters used in the experiments using grid search on a validation subset of the training data. The filter wise parameters used based on the result of parameter tuning in the experiment are given below:

    **i.** Mean filter (cv2.blur):
- Parameters: Kernel size = (5,5) ; Chosen the kernel size with highest PSNR.

    **ii.** Gaussian Filter (cv2.GaussianBlur):
- Parameters: Kernel size = (5,5) ;
  σ = 0 (auto-computed by OpenCV)

    **iii.** Median Filter (cv2.medianBlur):
- Parameters: Kernel size = (5,5) ; Chosen the kernel size with highest PSNR.

    **iv.** Bilateral filter (cv2.bilateralFilter):
- Parameters: Diameter (d) = 9
  σColor = 75
  σSpace = 75

    **v.** BM3D filter (cv2.fastNlMeansDenoisingColored):
- Parameters: h = 10 (luma filtering strength)
  hColor = 10 (chrominance filtering strength)
  templateWindowSize = 7
  searchWindowSize = 21

*Effectiveness of the experiments enhancing rice leaf images with Gaussian Noise*

*Table 1. Performance of mean filter denoising images with Gaussian Noise*

| Metric | Denoise | CD (2.0,8) | CD (2.0,5) | CD (1.0,5) | CD (0.5,5) | DC (2.0,8) | DC (2.0,5) | DC (1.0,5) | DC (0.5,5) |
|---|---|---|---|---|---|---|---|---|---|
| MSE | 67.3 | 1150.18 | 1221.92 | 363.77 | 152.32 | 242.32 | 284.44 | 133.63 | **61.18** |
| SSIM | 0.82 | 0.74 | 0.74 | 0.79 | 0.81 | 0.72 | 0.72 | 0.78 | **0.89** |
| PSNR | 30.04 | 17.59 | 17.34 | 22.58 | 26.37 | 24.37 | 23.7 | 26.97 | **30.43** |
| NRMSE | 0.04 | 0.16 | 0.17 | 0.09 | 0.06 | 0.08 | 0.08 | 0.06 | **0.04** |
| NMI | 1.22 | 1.14 | 1.15 | 1.18 | 1.2 | 1.14 | 1.14 | 1.17 | **1.25** |

*Table 2. Performance of Gaussian filter denoising images with Gaussian Noise*

| Metric | Denoise | CD (2.0,8) | CD (2.0,5) | CD (1.0,5) | CD (0.5,5) | DC (2.0,8) | DC (2.0,5) | DC (1.0,5) | DC (0.5,5) |
|---|---|---|---|---|---|---|---|---|---|
| MSE | 69.7 | 1179.37 | 1249.17 | 374.68 | 158.32 | 345.93 | 367.12 | 163.16 | **51.05** |
| SSIM | 0.75 | 0.66 | 0.66 | 0.72 | 0.74 | 0.64 | 0.64 | 0.71 | **0.87** |
| PSNR | 29.78 | 17.48 | 17.23 | 22.44 | 26.18 | 22.78 | 22.56 | 26.06 | **31.17** |
| NRMSE | 0.04 | 0.17 | 0.17 | 0.09 | 0.06 | 0.09 | 0.09 | 0.06 | **0.03** |
| NMI | 1.19 | 1.12 | 1.13 | 1.16 | 1.18 | 1.12 | 1.13 | 1.15 | **1.24** |

*Table 3. Performance of Median filter denoising images with Gaussian Noise*

| Metric | Denoise | CD (2.0,8) | CD (2.0,5) | CD (1.0,5) | CD (0.5,5) | DC (2.0,8) | DC (2.0,5) | DC (1.0,5) | DC (0.5,5) |
|---|---|---|---|---|---|---|---|---|---|
| MSE | 60.63 | 1080.69 | 1160.3 | 328.44 | 131.81 | 336.81 | 367.12 | 155.79 | **52.92** |
| SSIM | 0.75 | 0.64 | 0.64 | 0.71 | 0.74 | 0.63 | 0.63 | 0.71 | **0.84** |
| PSNR | 30.39 | 17.87 | 17.57 | 23.03 | 26.98 | 22.92 | 22.59 | 26.27 | **30.99** |
| NRMSE | 0.04 | 0.16 | 0.17 | 0.09 | 0.06 | 0.09 | 0.09 | 0.06 | **0.04** |
| NMI | 1.2 | 1.13 | 1.14 | 1.17 | 1.19 | 1.12 | 1.13 | 1.16 | **1.22** |

*Table 4. Performance of Bilateral filter denoising images with Gaussian Noise*

| Metric | Denoise | CD (2.0,8) | CD (2.0,5) | CD (1.0,5) | CD (0.5,5) | DC (2.0,8) | DC (2.0,5) | DC (1.0,5) | DC (0.5,5) |
|---|---|---|---|---|---|---|---|---|---|
| MSE | 53.66 | 1467.79 | 1508.11 | 389.5 | 146.02 | 255.07 | 279.72 | 120.89 | **29.69** |
| SSIM | 0.77 | 0.37 | 0.37 | 0.61 | 0.72 | 0.63 | 0.63 | 0.72 | **0.92** |
| PSNR | 30.86 | 16.51 | 16.4 | 22.26 | 26.51 | 24.11 | 23.72 | 27.35 | **33.56** |
| NRMSE | 0.04 | 0.19 | 0.19 | 0.1 | 0.06 | 0.08 | 0.08 | 0.05 | **0.03** |
| NMI | 1.21 | 1.09 | 1.1 | 1.15 | 1.18 | 1.14 | 1.14 | 1.17 | **1.29** |

*Table 5. Performance of BM3D filter denoising images with Gaussian Noise*

| Metric | Denoise | CD (2.0,8) | CD (2.0,5) | CD (1.0,5) | CD (0.5,5) | DC (2.0,8) | DC (2.0,5) | DC (1.0,5) | DC (0.5,5) |
|---|---|---|---|---|---|---|---|---|---|
| MSE | 43.09 | 1842.22 | 1831.14 | 399.42 | 144.87 | 122.03 | 118.62 | 74.47 | **29.67** |
| SSIM | 0.91 | 0.4 | 0.41 | 0.85 | 0.9 | 0.87 | 0.87 | 0.89 | **0.92** |
| PSNR | 32.06 | 15.55 | 15.59 | 22.14 | 26.57 | 27.91 | 27.78 | 29.71 | **33.53** |
| NRMSE | 0.03 | 0.21 | 0.21 | 0.1 | 0.06 | 0.05 | 0.05 | 0.04 | **0.03** |
| NMI | 1.29 | 1.08 | 1.09 | 1.21 | 1.26 | 1.21 | 1.2 | 1.23 | **1.3** |

From Table 1, Table 2, Table 3, Table 4, and Table 5, we have observed that in the experiment 'Exp.09: Denoise-CLAHE (0.5,5) performed well with all the filters. Denoising with Bilateral filter or BM3D filter, followed by application of CLAHE with Clip Limit=0.5 and Tile Grid Size=(5,5), was most effective in enhancing an image of rice leaf of size 256x256 with Gaussian noise. One of the noisy image with Gaussian noise and its enhanced images are shown in the Fig.6. (a), (b), and (c) respectively.

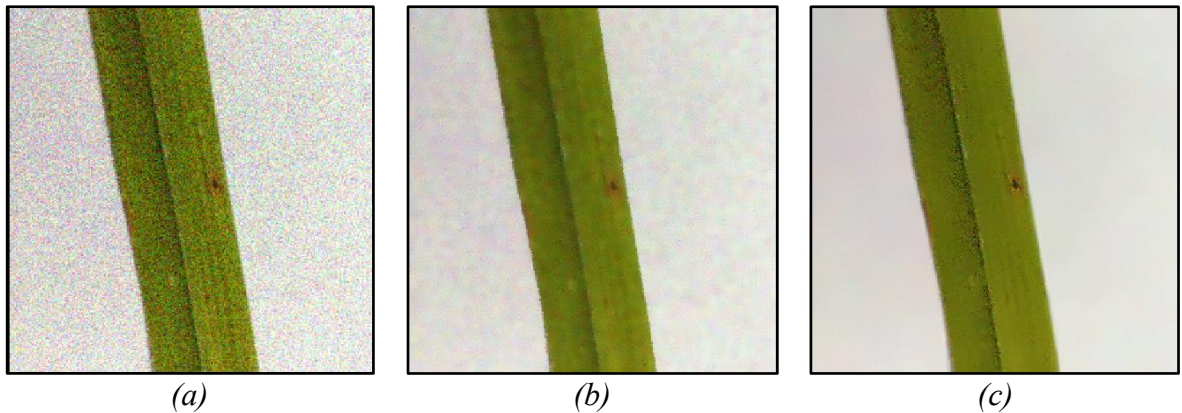

*(a)* *(b)* *(c)*

***Fig.6.*** *(a) Image with Gaussian Noise, (b) Image after application of CLAHE (0.5,5) to denoised image with Bilateral filter, (c) Image after application of CLAHE (0.5,5) to denoised image with BM3D filter*

***Effectiveness of the experiments enhancing rice leaf images with Salt-and-pepper Noise***

*Table 6. Performance of mean filter denoising images with Salt-and-pepper Noise*

| Metric | Denoise | CD (2.0,8) | CD (2.0,5) | CD (1.0,5) | CD (0.5,5) | DC (2.0,8) | DC (2.0,5) | DC (1.0,5) | DC (0.5,5) |
|---|---|---|---|---|---|---|---|---|---|
| MSE | **63.53** | 181.92 | 214.47 | 107.32 | 79.72 | 234.1 | 300.5 | 148.76 | 97.6 |
| SSIM | **0.88** | 0.88 | 0.88 | 0.88 | 0.88 | 0.78 | 0.76 | 0.82 | 0.85 |
| PSNR | **30.31** | 25.87 | 25.09 | 28.05 | 29.27 | 24.56 | 23.5 | 26.5 | 28.32 |
| NRMSE | **0.04** | 0.06 | 0.07 | 0.05 | 0.04 | 0.07 | 0.08 | 0.06 | 0.05 |
| NMI | **1.26** | 1.18 | 1.18 | 1.21 | 1.23 | 1.16 | 1.17 | 1.2 | 1.22 |

*Table 7. Performance of Gaussian filter denoising images with Salt-and-pepper Noise*

| Metric | Denoise | CD (2.0,8) | CD (2.0,5) | CD (1.0,5) | CD (0.5,5) | DC (2.0,8) | DC (2.0,5) | DC (1.0,5) | DC (0.5,5) |
|---|---|---|---|---|---|---|---|---|---|
| MSE | **65.32** | 184.58 | 216.29 | 108.98 | 81.48 | 283.04 | 328.84 | 174.75 | 103.53 |
| SSIM | **0.87** | 0.87 | 0.87 | 0.87 | 0.87 | 0.78 | 0.78 | 0.81 | 0.84 |
| PSNR | **30.08** | 25.76 | 25.04 | 27.94 | 29.11 | 23.69 | 23.08 | 25.78 | 28.03 |
| NRMSE | **0.04** | 0.07 | 0.07 | 0.05 | 0.04 | 0.08 | 0.09 | 0.06 | 0.05 |
| NMI | **1.29** | 1.19 | 1.19 | 1.23 | 1.26 | 1.17 | 1.17 | 1.22 | 1.25 |

Chutia et al.: Enhancing rice leaf images: An overview of image denoising techniques

*Table 8. Performance of Median filter denoising images with Salt-and-pepper Noise*

| Metric | Denoise | CD (2.0,8) | CD (2.0,5) | CD (1.0,5) | CD (0.5,5) | DC (2.0,8) | DC (2.0,5) | DC (1.0,5) | DC (0.5,5) |
|---|---|---|---|---|---|---|---|---|---|
| MSE | **15.24** | 116.95 | 140.77 | 50.73 | 28.01 | 81.51 | 98.25 | 46.01 | **27.81** |
| SSIM | **0.94** | 0.92 | 0.92 | 0.93 | 0.94 | 0.92 | 0.92 | 0.93 | **0.93** |
| PSNR | **37.16** | 27.97 | 27.08 | 31.57 | 34.02 | 29.41 | 28.67 | 31.89 | **34.07** |
| NRMSE | **0.02** | 0.05 | 0.06 | 0.03 | 0.03 | 0.04 | 0.05 | 0.03 | **0.03** |
| NMI | **1.37** | 1.22 | 1.22 | 1.29 | 1.32 | 1.24 | 1.23 | 1.29 | **1.32** |

*Table 9. Performance of Bilateral filter denoising images with Salt-and-pepper Noise*

| Metric | Denoise | CD (2.0,8) | CD (2.0,5) | CD (1.0,5) | CD (0.5,5) | DC (2.0,8) | DC (2.0,5) | DC (1.0,5) | DC (0.5,5) |
|---|---|---|---|---|---|---|---|---|---|
| MSE | 511.23 | 584.37 | 606.67 | 527.08 | 511.22 | 560.28 | 567.39 | 523.81 | 511.95 |
| SSIM | 0.86 | 0.88 | 0.87 | 0.87 | 0.86 | 0.85 | 0.85 | 0.86 | 0.86 |
| PSNR | 21.07 | 20.51 | 20.36 | 20.95 | 21.07 | 20.67 | 20.63 | 20.97 | 21.07 |
| NRMSE | 0.11 | 0.12 | 0.12 | 0.11 | 0.11 | 0.11 | 0.12 | 0.11 | 0.11 |
| NMI | 1.36 | 1.24 | 1.24 | 1.29 | 1.32 | 1.24 | 1.23 | 1.28 | 1.32 |

*Table 10. Performance of BM3D filter denoising images with Salt-and-pepper Noise*

| Metric | Denoise | CD (2.0,8) | CD (2.0,5) | CD (1.0,5) | CD (0.5,5) | DC (2.0,8) | DC (2.0,5) | DC (1.0,5) | DC (0.5,5) |
|---|---|---|---|---|---|---|---|---|---|
| MSE | 518.25 | 606.88 | 630.4 | 541.3 | 521.66 | 558.89 | 550.62 | 528.13 | 518.52 |
| SSIM | 0.86 | 0.83 | 0.83 | 0.85 | 0.85 | 0.84 | 0.84 | 0.85 | 0.85 |
| PSNR | 21.01 | 20.35 | 20.19 | 20.83 | 20.98 | 20.67 | 20.75 | 20.93 | 21.01 |
| NRMSE | 0.11 | 0.12 | 0.12 | 0.11 | 0.11 | 0.11 | 0.11 | 0.11 | 0.11 |
| NMI | 1.35 | 1.22 | 1.22 | 1.28 | 1.31 | 1.25 | 1.24 | 1.28 | 1.31 |

From Table 6, Table 7, Table 8, Table 9, and Table 10, we have observed that in the experiment 'Exp.01: Denoise' performed best as compared to the rest of the experiments with all the filters in enhancing image with Salt-and-pepper noise. Denoising with Median filter was most effective in enhancing an image of rice leaf of size 256x256 with Salt-and-pepper noise.

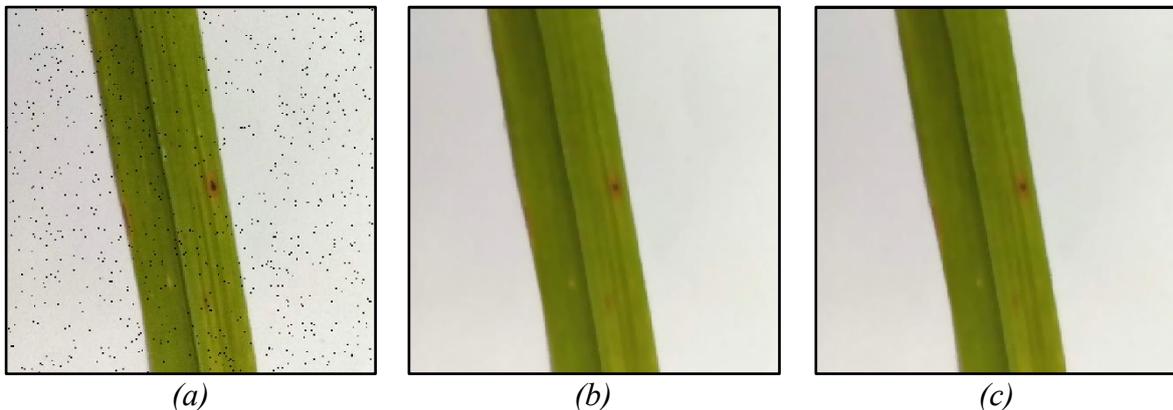

*(a)*            *(b)*            *(c)*

**Fig.7.** *(a) Image with Salt-and-pepper Noise, (b) Image after denoising with Median filter, (c) Image after application of CLAHE (0.5,5) to the denoised image with Median filter*

One of the noisy images and its enhanced images are shown in the Fig.7. (a), (b), and (c) respectively. We have seen from the experiments that application of CLAHE with Clip Limit=0.5 and Tile Grid Size = (5,5) to the denoised images with Median filter enhances the details in the image.

*Effectiveness of the experiments enhancing rice leaf images with Speckle Noise*

*Table 11. Performance of mean filter denoising images with Speckle Noise*

| Metric | Denoise | CD (2.0,8) | CD (2.0,5) | CD (1.0,5) | CD (0.5,5) | DC (2.0,8) | DC (2.0,5) | DC (1.0,5) | DC (0.5,5) |
|---|---|---|---|---|---|---|---|---|---|
| MSE | **57.85** | 1054.19 | 1073.65 | 347.68 | 146.66 | 213.13 | 259.75 | 122.5 | 86.55 |
| SSIM | **0.84** | 0.77 | 0.77 | 0.81 | 0.83 | 0.75 | 0.75 | 0.8 | 0.82 |
| PSNR | **30.78** | 17.98 | 17.92 | 22.79 | 26.55 | 24.97 | 24.13 | 27.39 | 28.88 |
| NRMSE | **0.04** | 0.16 | 0.16 | 0.09 | 0.06 | 0.07 | 0.08 | 0.05 | 0.04 |
| NMI | **1.23** | 1.15 | 1.15 | 1.19 | 1.22 | 1.14 | 1.15 | 1.18 | 1.2 |

*Table 12. Performance of Gaussian filter denoising images with Speckle Noise*

| Metric | Denoise | CD (2.0,8) | CD (2.0,5) | CD (1.0,5) | CD (0.5,5) | DC (2.0,8) | DC (2.0,5) | DC (1.0,5) | DC (0.5,5) |
|---|---|---|---|---|---|---|---|---|---|
| MSE | **54.59** | 1072.74 | 1090.95 | 351.88 | 146.59 | 285.17 | 315.29 | 138.49 | 89.95 |
| SSIM | **0.79** | 0.7 | 0.69 | 0.75 | 0.77 | 0.68 | 0.68 | 0.74 | 0.77 |
| PSNR | **30.91** | 17.91 | 17.84 | 22.74 | 26.53 | 23.68 | 23.28 | 26.84 | 28.68 |
| NRMSE | **0.04** | 0.16 | 0.16 | 0.09 | 0.06 | 0.08 | 0.09 | 0.06 | 0.05 |
| NMI | **1.22** | 1.14 | 1.14 | 1.18 | 1.2 | 1.13 | 1.14 | 1.17 | 1.19 |

*Table 13. Performance of Median filter denoising images with Speckle Noise*

| Metric | Denoise | CD (2.0,8) | CD (2.0,5) | CD (1.0,5) | CD (0.5,5) | DC (2.0,8) | DC (2.0,5) | DC (1.0,5) | DC (0.5,5) |
|---|---|---|---|---|---|---|---|---|---|
| MSE | **45.01** | 998.17 | 1021.96 | 314.24 | 124.25 | 276.65 | 319.21 | 131.02 | 78.62 |
| SSIM | **0.79** | 0.68 | 0.67 | 0.75 | 0.77 | 0.67 | 0.67 | 0.74 | 0.77 |
| PSNR | **31.73** | 18.23 | 18.13 | 23.23 | 27.25 | 23.86 | 23.27 | 27.1 | 29.26 |
| NRMSE | **0.03** | 0.15 | 0.16 | 0.09 | 0.05 | 0.08 | 0.09 | 0.06 | 0.04 |
| NMI | **1.22** | 1.14 | 1.15 | 1.19 | 1.21 | 1.13 | 1.14 | 1.17 | 1.19 |

*Table 14. Performance of Bilateral filter denoising images with Speckle Noise*

| Metric | Denoise | CD (2.0,8) | CD (2.0,5) | CD (1.0,5) | CD (0.5,5) | DC (2.0,8) | DC (2.0,5) | DC (1.0,5) | DC (0.5,5) |
|---|---|---|---|---|---|---|---|---|---|
| MSE | **35.07** | 1240.95 | 1245.15 | 346.28 | 125.95 | 175.58 | 220.51 | 88.67 | 57.57 |
| SSIM | **0.83** | 0.43 | 0.43 | 0.68 | 0.79 | 0.72 | 0.72 | 0.79 | 0.81 |
| PSNR | **32.77** | 17.26 | 17.25 | 22.8 | 27.18 | 25.8 | 24.84 | 28.82 | 30.62 |
| NRMSE | **0.03** | 0.17 | 0.17 | 0.09 | 0.05 | 0.06 | 0.07 | 0.05 | 0.04 |
| NMI | **1.24** | 1.11 | 1.12 | 1.17 | 1.21 | 1.16 | 1.16 | 1.2 | 1.22 |

*Table 15. Performance of BM3D filter denoising images with Speckle Noise*

| Metric | Denoise | CD (2.0,8) | CD (2.0,5) | CD (1.0,5) | CD (0.5,5) | DC (2.0,8) | DC (2.0,5) | DC (1.0,5) | DC (0.5,5) |
|---|---|---|---|---|---|---|---|---|---|
| MSE | **28.17** | 1283.43 | 1251.61 | 331.56 | 122.18 | 70.51 | 79.18 | 51.15 | 36.48 |
| SSIM | **0.91** | 0.54 | 0.57 | 0.9 | 0.91 | 0.89 | 0.89 | 0.9 | 0.91 |
| PSNR | **33.9** | 17.13 | 17.24 | 23.01 | 27.33 | 29.89 | 29.46 | 31.27 | 32.7 |
| NRMSE | **0.03** | 0.17 | 0.17 | 0.09 | 0.05 | 0.04 | 0.04 | 0.03 | 0.03 |
| NMI | **1.31** | 1.12 | 1.13 | 1.25 | 1.29 | 1.24 | 1.23 | 1.26 | 1.28 |

From Table 11, Table 12, Table 13, Table 14, and Table 15, we have observed that in the experiment 'Exp.01: Denoise' performed best as compared to the rest of the experiments with all the filters in enhancing image with Speckle noise. Denoising with BM3D filter was most effective followed by Bilateral filter in enhancing an image of rice leaf of size 256x256 with Speckle noise.

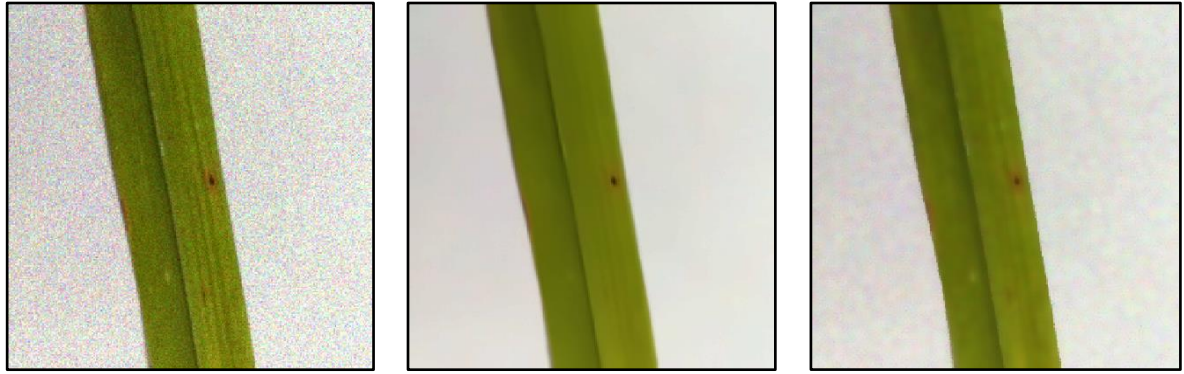

*(a)*           *(b)*           *(c)*

***Fig.8.** (a) Image with Speckle Noise, (b) Image after denoising with BM3D filter, (c) Image after denoising with Bilateral filter*

One of the noisy image and its enhanced images are shown in the ***Fig.8.*** *(a), (b),* and *(c)* respectively. We have visually observed from the result images that image denoised with BM3D filter has lesser amount of noise as compared to the image denoised with Bilateral filter, but the later one contains more details than the first one.

***Effectiveness of the experiments enhancing rice leaf images with Random Noise***

*Table 16. Performance of mean filter denoising images with Random Noise*

| Metric | Denoise | CD (2.0,8) | CD (2.0,5) | CD (1.0,5) | CD (0.5,5) | DC (2.0,8) | DC (2.0,5) | DC (1.0,5) | DC (0.5,5) |
|---|---|---|---|---|---|---|---|---|---|
| MSE | **42.75** | 383.28 | 531.95 | 160.84 | 67.64 | 131.05 | 167.18 | 88.4 | 61.18 |
| SSIM | **0.90** | 0.86 | 0.86 | 0.88 | 0.89 | 0.85 | 0.85 | 0.88 | 0.89 |
| PSNR | **32.29** | 22.49 | 21.01 | 26.19 | 30.02 | 27.14 | 26.08 | 28.81 | 30.43 |
| NRMSE | **0.03** | 0.09 | 0.11 | 0.06 | 0.04 | 0.06 | 0.06 | 0.05 | 0.04 |
| NMI | **1.29** | 1.18 | 1.18 | 1.23 | 1.26 | 1.18 | 1.18 | 1.22 | 1.25 |

*Table 17. Performance of Gaussian filter denoising images with Random Noise*

| Metric | Denoise | CD (2.0,8) | CD (2.0,5) | CD (1.0,5) | CD (0.5,5) | DC (2.0,8) | DC (2.0,5) | DC (1.0,5) | DC (0.5,5) |
|---|---|---|---|---|---|---|---|---|---|
| MSE | **32.12** | 382.72 | 531.04 | 153.53 | 58.39 | 142 | 182.91 | 78.79 | 51.05 |
| SSIM | **0.89** | 0.83 | 0.83 | 0.87 | 0.88 | 0.82 | 0.82 | 0.86 | 0.87 |
| PSNR | **33.42** | 22.5 | 21.01 | 26.39 | 30.61 | 26.77 | 25.67 | 29.31 | 31.17 |
| NRMSE | **0.03** | 0.09 | 0.11 | 0.06 | 0.04 | 0.06 | 0.07 | 0.04 | 0.03 |
| NMI | **1.28** | 1.17 | 1.17 | 1.22 | 1.25 | 1.17 | 1.17 | 1.22 | 1.24 |

*Table 18. Performance of Median filter denoising images with Random Noise*

| Metric | Denoise | CD (2.0,8) | CD (2.0,5) | CD (1.0,5) | CD (0.5,5) | DC (2.0,8) | DC (2.0,5) | DC (1.0,5) | DC (0.5,5) |
|---|---|---|---|---|---|---|---|---|---|
| MSE | **32.13** | 380.44 | 533.03 | 154.05 | 58.71 | 169.37 | 214.43 | 85.11 | 52.92 |
| SSIM | **0.85** | 0.79 | 0.78 | 0.83 | 0.84 | 0.76 | 0.76 | 0.82 | 0.84 |
| PSNR | **33.26** | 22.51 | 20.99 | 26.38 | 30.56 | 25.98 | 24.98 | 28.98 | 30.99 |
| NRMSE | **0.03** | 0.09 | 0.11 | 0.06 | 0.04 | 0.06 | 0.07 | 0.04 | 0.04 |
| NMI | **1.25** | 1.16 | 1.17 | 1.21 | 1.23 | 1.15 | 1.16 | 1.20 | 1.22 |

*Table 19. Performance of Bilateral filter denoising images with Random Noise*

| Metric | Denoise | CD (2.0,8) | CD (2.0,5) | CD (1.0,5) | CD (0.5,5) | DC (2.0,8) | DC (2.0,5) | DC (1.0,5) | DC (0.5,5) |
|---|---|---|---|---|---|---|---|---|---|
| MSE | **18.42** | 390.18 | 541.21 | 135.81 | 42.33 | 83.48 | 112.52 | 49.83 | **29.69** |
| SSIM | **0.92** | 0.74 | 0.73 | 0.90 | 0.92 | 0.90 | 0.90 | 0.91 | **0.92** |
| PSNR | **35.68** | 22.38 | 20.92 | 26.96 | 32.03 | 29.26 | 27.96 | 31.43 | **33.56** |
| NRMSE | **0.02** | 0.10 | 0.11 | 0.06 | 0.03 | 0.04 | 0.05 | 0.03 | **0.03** |
| NMI | **1.32** | 1.15 | 1.16 | 1.24 | 1.29 | 1.21 | 1.21 | 1.26 | **1.29** |

*Table 20. Performance of BM3D filter denoising images with Random Noise*

| Metric | Denoise | CD (2.0,8) | CD (2.0,5) | CD (1.0,5) | CD (0.5,5) | DC (2.0,8) | DC (2.0,5) | DC (1.0,5) | DC (0.5,5) |
|---|---|---|---|---|---|---|---|---|---|
| MSE | **19.84** | 381.66 | 534.36 | 140.46 | 45.24 | 72.35 | 63.89 | 40.79 | **29.67** |
| SSIM | **0.92** | 0.91 | 0.91 | 0.92 | 0.92 | 0.91 | 0.91 | 0.92 | **0.92** |
| PSNR | **35.37** | 22.54 | 20.99 | 26.82 | 31.78 | 29.82 | 30.43 | 32.24 | **33.53** |
| NRMSE | **0.02** | 0.09 | 0.11 | 0.06 | 0.03 | 0.04 | 0.04 | 0.03 | **0.03** |
| NMI | **1.32** | 1.22 | 1.21 | 1.27 | 1.30 | 1.25 | 1.25 | 1.27 | **1.30** |

From Table 16, Table 17, Table 18, Table 19, and Table 20, we have observed that in the experiment 'Exp.01: Denoise' performed best as compared to the rest of the experiments with all the filters in enhancing image with Random noise. Denoising with Bilateral filter was most effective followed by BM3D filter in enhancing an image of rice leaf of size 256x256 with Random noise.

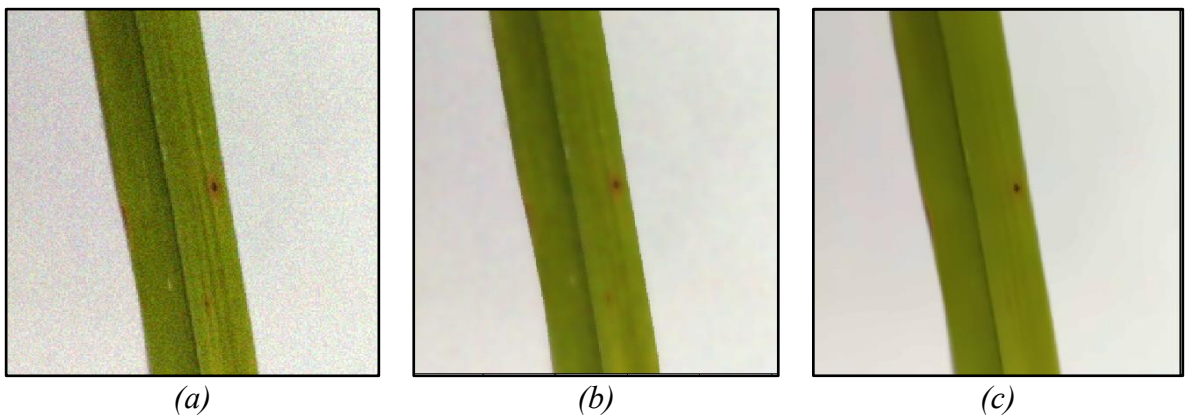

*(a)* *(b)* *(c)*

**Fig.9.** *(a) Image with Random Noise, (b) Image after denoising with Bilateral filter, (c) Image after denoising with BM3D filter*

One of the noisy image and its enhanced images are shown in the **Fig.9.** *(a), (b),* and *(c)* respectively. We have visually observed from the result images that image denoised with Bilateral filter contain more details than the image denoised with BM3D. The key findings of the experiments is summarized in **Table 21**.

*Table 21. Key finding of the study*

| Noise Type | Best denoising filter(s) | CLAHE enhancement |
|---|---|---|
| **Gaussian noise** | Bilateral or BM3D | Clip Limit=0.5, Tile Grid Size=(5,5) |
| **Salt-and-pepper noise** | Median | Clip Limit=0.5, Tile Grid Size=(5,5) |
| **Speckle Noise** | BM3D followed by Bilateral | - |
| **Random Noise** | Bilateral followed by BM3D | - |

*Time analysis of the experiments:*

There is a variation in time required for denoising the images with different types of noise by different filters. Therefore, we have recorded the time consumed by each filter for denoising 500 noisy images of rice leaves, which is given in the ***Table 22***. The time required by the filters depend on various factors, mainly the size of the image, which in our case defined as 256x256.

*Table 22. Time required by different filters for denoising 500 images*

| Filter | Time required for denoising 500 images (in seconds) |
|---|---|
| **Mean Filter** | 0.08-0.39 |
| **Gaussian Filter** | 0.08-0.39 |
| **Median Filter** | 0.08-0.39 |
| Bilateral Filter | 4.0-7.0 |
| BM3D Filter | 110.0-140.0 |

**CONCLUSION:**

The study was conducted for evaluating the efficiency of different image filters combined with CLAHE (Contrast Limited Adaptive Histogram Equalization) for enhancing the images of rice leaves. Among the filters used in the experiments, Median, Bilateral, and BM3D filter performed very well with or without CLAHE enhancement. From the above results we can conclude with the following points:

a) When Bilateral or BM3D denoising combined with CLAHE (0.5,(5,5)) applied to images containing Gaussian noise, the resulting images show the clearest leaf structures and highest contrast.
b) Salt-and-pepper noise is best removed using the Median filter; veins and lesion boundaries may be brought out afterward by applying CLAHE (0.5,(5,5)).
c) BM3D reduces persistent speckle to the lowest degree; meanwhile, the Bilateral filter more effectively preserves fine edges depending on one's priorities for noise suppression or edge detail.
d) Bilateral filter maintains the smoothing and edge preservation while denoising image with random noise.
e) Applying CLAHE with a high clip limit to noisy images can lead to more prominent noise pixels, and the critical details may be lost. Therefore, we can apply a lower clip limit with optimal tile size on the denoised image in order to enhance details in the image.
f) BM3D filter consumes a significant amount of time for denoising image as compared to other filters, therefore, it is not suggested to use BM3D filter to process huge dataset.

Median filter is one of fastest non-linear filter and as fast as linear filters like Mean and Gaussian in processing images of 256x256 size. By compromising a little bit in the speed, Bilateral filter provides very good result in denoising images with Gaussian, speckle and random noise.

The study results will also help to develop more accurate and intelligent image analysis systems for diagnosing rice leaf diseases and detecting nutrient deficiencies. The results can offer great help for researchers and practitioners in the field of agricultural image processing in choosing most appropriate image denoising methods for their applications. On the other hand, the study will contribute to improved crop management and productivity through more accurate disease diagnosis and nutrient deficiency determination.

**REFERENCES**

[1] Speranskyy, V. and Balaban, D. S. (2024): Analysis of methods and algorithms for image filtering and quality enhancement. Applied Aspects of Information Technology 7(3): 255–268. https://doi.org/10.15276/aait.07.2024.18

[2] Verma, R. and Ali, J. (2013): A comparative study of various types of image noise and efficient noise removal techniques. International Journal of Advanced Research in Computer Science and Software Engineering 3(10).

[3] Jahan (2017): A comparative study of Gaussian noise removal methodologies for gray scale images. International Journal of Computer Applications 172: 1–6.

[4] Dong, F., Chen, Y., Kong, D. X. and Yang, B. (2015): Salt and pepper noise removal based on an approximation of l0 norm. Computers & Mathematics with Applications 70(5): 789–804.

[5] McHugh, S. (2025): Digital camera image noise: concept and types. Cambridge in Colour. Available online: https://www.cambridgeincolour.com/tutorials/image-noise.htm (Accessed: Mar. 8, 2025).

[6] Naser, A. (2024): A proposed CLCOA technique based on CLAHE using Cat Optimized Algorithm for plants images enhancement. Wasit Journal of Computer and Mathematical Sciences. https://doi.org/10.31185/wjcms.202

[7] Sun, Y., Wang, Y. and Yang, Q. (2024): Welding image enhancement based on CLAHE and guided filter. In: Proceedings of the 2024 International Conference on Electrical, Electronics and Computer Research (EECR), pp. 285–290. https://doi.org/10.1109/eecr60807.2024.10607276

[8] Wilson, B. (2023): Histogram equalization – Part 1 – Theory. Ben Wilson Blog. Available online: https://benqwilson.com/blog/R8Vz/histogram-equalization-part-1-theory (Accessed: Sep. 17, 2023).

[9] Kuran, U. and Kuran, E. C. (2021): Parameter selection for CLAHE using multi-objective cuckoo search algorithm for image contrast enhancement. Intelligent Systems with Applications 12: 200051. https://doi.org/10.1016/j.iswa.2021.200051

[10] Deng, G. and Cahill, L. W. (1993): An adaptive Gaussian filter for noise reduction and edge detection. IEEE Nuclear Science Symposium and Medical Imaging Conference, pp. 1615–1619. https://doi.org/10.1109/NSSMIC.1993.373563

[11] Ai, H., Nguyen, V. L., Luong, K. T. and Le, V. T. (2024): Design of mean filter using field programmable gate arrays for digital images. Indonesian Journal of Electrical Engineering and Computer Science 36(3): 1430–1436. https://doi.org/10.11591/ijeecs.v36.i3.pp1430-1436

[12] Dinç, İ., et al. (2015): DT-Binarize: A decision tree based binarization for protein crystal images. In: Deligiannidis, L. and Deligiannidis, H. R. L. (eds) Emerging Trends in Image Processing, Computer Vision and Pattern Recognition. pp. 183–199. https://doi.org/10.1016/B978-0-12-802045-6.00012-0

[13] Gong, Y. (2023): Hardware oriented bilateral filter algorithm and architecture survey. Applied Computing in Engineering 2: 490–498.

[14] ShaR (2025): Rice Leafs – An image collection of four rice diseases. Kaggle. Available online: https://www.kaggle.com/datasets/shayanriyaz/riceleafs/data

[15] Scikit-learn developers (2025): sklearn.metrics.normalized_mutual_info_score. Scikit-learn Documentation. Available online: https://scikit-learn.org/stable/modules/generated/sklearn.metrics.normalized_mutual_info_score.html (Accessed: Jun. 25, 2025).